# PERFORMANCE ANALYSIS OF REGULARIZED LINEAR REGRESSION MODELS FOR OXAZOLINES AND OXAZOLES DERIVATIVES DESCRIPTOR DATASET

Doreswamy[1] and Chanabasayya .M. Vastrad[2]

[1]Department of Computer Science Mangalore University , Mangalagangotri-574 199, Karnataka, INDIA
[2]Department of Computer Science Mangalore University , Mangalagangotri-574 199, Karnataka, INDIA

## ABSTRACT

*Regularized regression techniques for linear regression have been created the last few ten years to reduce the flaws of ordinary least squares regression with regard to prediction accuracy. In this paper, new methods for using regularized regression in model choice are introduced, and we distinguish the conditions in which regularized regression develops our ability to discriminate models. We applied all the five methods that use penalty-based (regularization) shrinkage to handle Oxazolines and Oxazoles derivatives descriptor dataset with far more predictors than observations. The lasso, ridge, elasticnet, lars and relaxed lasso further possess the desirable property that they simultaneously select relevant predictive descriptors and optimally estimate their effects. Here, we comparatively evaluate the performance of five regularized linear regression methods The assessment of the performance of each model by means of benchmark experiments is an established exercise. Cross-validation and resampling methods are generally used to arrive point evaluates the efficiencies which are compared to recognize methods with acceptable features. Predictive accuracy was evaluated using the root mean squared error (RMSE) and Square of usual correlation between predictors and observed mean inhibitory concentration of antitubercular activity (R square). We found that all five regularized regression models were able to produce feasible models and efficient capturing the linearity in the data. The elastic net and lars had similar accuracies as well as lasso and relaxed lasso had similar accuracies but outperformed ridge regression in terms of the RMSE and R square metrics.*

## KEYWORDS

## 1. INTRODUCTION

Multiple linear regression is frequently employ to evaluate a model for predicting expected responses, or to investigate the relationship between the response (activity) and the predictors(molecular descriptors) . For the first aim the prediction accuracy of the model is important, for the second aim the complexity of the model is of more interest. Common linear regression procedures are popular for often not carrying out well with respect to both prediction performance and model involvement. Regularization plays a key role in the analysis of modern data. Regularized regression methods for linear regression have been evolved to beat the defects of ordinary least squares regression with regard to prediction accuracy. Regularization [2] has been deeply studied on the interface between statistics and computer science. The research on regularized results has gained increasing interest during the last ten years [1]. This is in some ways due to advances in measurement technologies, e.g., in molecular pharmaceutics, where high-throughput technologies allow simultaneous measurement of tens of hundreds of predictor variables(molecular descriptors). However, the measurements are valuable, so typically the number of data points is small. In the field of drug development, the number of descriptors is not that large but yet enough to prevent the use of many standard data analysis methods. Conventional regression methods are unable to process data with more predictors than observations(so called $p \gg n$ problem). Regularization methods [3] help in formalizing a unique solution in this well posed problem. These methods shrink some of the coefficients to zero. This not alone helps in descriptor selection but in addition reduces the difference at the cost of a small advance in bias. However, this has the outcome of improving the generalization of the estimate.

The prediction of antituberculer activity using Oxazolines and Oxazoles derivatives is currently undertaken using various regularization methods with varying amount of complexity, computational power and predictive accuracy. Performance evaluation of existing methods are thus essential to identify those best suited to prediction and decide when their performance is optimal. Here, we evaluate the relative performance of five regularized linear regression models for prediction. The methods comprise the Least absolute shrinkage and selection operator (Lasso)[4-5], Ridge regression (RR) [5-7], Elastic net[5,8-10],Relaxed lasso[11-12],Least Angle Regression(LARS)[13-14]. The claim and arrival of regularization models in various application fields, containing descriptor selection, associated to their use of penalties that eases fitting models with variables that run towards thousands, including many irrelevant to the response, far exceed the sample size, or are highly correlated, with high efficiency and prediction accuracy.

## 2. MATERIALS AND METHODS

### 2.1 The Molecular Descriptor Data Set

The molecular descriptors of hundred Oxazolines and Oxazoles derivatives [15-16] based H37Rv inhibitors analyzed. These molecular descriptors are produced using Padel-Descriptor tool [17]. The dataset includes a various set of molecular descriptors with a broad range of inhibitory activities versus H37Rv. This molecular descriptor data set covers 100 observations with 234 descriptors. Before modelling, the dataset is centred.

### 2.2 The regularization models

The fundamental linear regression method used to predict antituberculer activity using Oxazolines and Oxazoles derivatives with all the five regularization methods is :

$$y = \mu 1_n + X\beta + e \qquad (1)$$

where $y = (y_1, \ldots, y_n)^T$ is the vector of observed antitubercular activity i.e mean inhibitory concentration (MIC), $1_n$ is a column vector of $n$ ones and $\mu$ is a common intercept, $X$ is a $n \times p$ matrix of molecular descriptors ; $\beta$ is the vector of the regression coefficients of the molecular descriptors and $e$ is the vector of the residual errors with var $(e) = I\sigma_e^2$ In what follows, we believe that the observed antitubercular activity have been mean-centred.

## 2.3 Lasso

Lasso regression techniques are broadly used in fields with large datasets, such as drug discovery, where effective and speedy algorithms are required[5], the lasso is also known as basis pursuit[18]. The lasso is, still, not robust to steep correlations between descriptors and will immediately select one and reject the others and cut down when all descriptors are alike. The lasso penalty look for m several coefficients to be near to zero, and only a modest subset to be best(and nonzero). The lasso estimator [4] uses the $\ell_1$ penalized least squares basis to obtain a sparse solution to the following optimization problem.

$$\hat{\beta}(lasso) = arg \min_{\beta}\|y - X\beta\|_2^2 + \lambda\|\beta\|_1 \qquad (2)$$

where $\|\beta\|_1 = \sum_{j=1}^{p}|\beta_j|$ is the $\ell_1$-norm penalty on $\beta$ , which activates sparsity in the solution, and $\lambda \geq 0$ is a tuning parameter.

The $\ell_1$ penalty allows the lasso to concurrently regularize the least squares fit and reduces few components of $\hat{\beta}(lasso)$ to zero for some well selected $\lambda$. The cyclical coordinate descent algorithm [5] smoothly computes the entire lasso solution paths for $\lambda$ the lasso technique and is faster than the popular LARS method[13-14]. These properties make the lasso an attractive and most popular predictor selection method. After all, the lasso has three important week points– it deficits the oracle property (see below), is weak with multi-dimensional data and can not choose large predictors than the sample size earlier it saturates when $p > n$ [19]. An oracle method can measure the subset of valid parameters with zero coefficients as definitely zero with probability likely to 1; that is, as well as if the true subset model were known ahead[20]. An oracle estimator, furthermore, performs asymptotically consistent and efficient predictor selection and produces asymptotically fair and normally distributed estimates of the nonzero coefficients[19-20]. The oracle feature is nearly related to the super-performance event[20]. Optimal estimators must also meet valid additional and key regularity settings as well having the oracle feature, such as continuous shrinkage [19]. The lasso deficits the oracle feature as it evaluates the extensive nonzero coefficients with asymptotically non-recognizable bias [20] and can only usually carry out predictor selection when the predictor matrix(or the descriptor matrix) fulfils a moderately powerful condition [19].

## 2.4 Ridge regression

Ridge regression [6] is perfect if there are various predictors, all with non-zero coefficients and collect from a normal distribution [5]. In specific, it carry out well with many predictors each having small outcome and prevents coefficients of linear regression models with many correlated predictors from being poorly determined and exhibiting high variance. Ridge regression shrinks the coefficients of correlated predictors equally towards zero. So, e.g., given $k$ alike predictors, each would get alike coefficients according to $1/k$th the extent that any one predictor would get if fit individually [5]. Ridge regression thus does not impose coefficients to disappear and therefore cannot choose a model with only the practically appropriate and predictive subset of variables.

The ridge regression estimator deals the regression problem in Eqn. (1) using $\ell_2$ penalized least squares:

$$\hat{\beta}(ridge) = arg \min_{\beta} \|y - X\beta\|_2^2 + \lambda \|\beta\|_2^2 \qquad (3)$$

where $\|y - X\beta\|_2^2 = \sum_{i=1}^{n}(y_i - x_i^T\beta)^2$ is the $\ell_2$-norm (quadratic) loss function (i.e. residual sum of squares), $x_i^T$ is the i-th row of X, $\|\beta\|_2^2 = \sum_{j=1}^{p}\beta_j^2$ is the $\ell_2$- norm penalty on $\beta$ and $\lambda \geq 0$ is the tuning parameter (penalty, regularization) which regulates the power of the penalty (linear shrinkage) by deciding the relative importance of the data-dependent practical error and the penalty term. The bigger the value of $\lambda$, the larger is the amount of shrinkage. As the value of $\lambda$ is reliant on the data it can be found out using data-possessed methods, such as cross-validation. The intercept is expected to be zero in Eqn. (2) due to mean centring of the molecular descriptor dataset.

## 2.5 Elastic net

The is an continuance of the lasso that is robust to highest correlations among the predictors [5]. To avoid the imbalance of the lasso solution paths when predictors are highly correlated , the elastic net was projected for evaluating high dimensional data [8]. The elastic net uses a mixture of the $\ell_1$ (lasso) and $\ell_2$ (ridge) penalties and can be defined as:

$$\hat{\beta}(elasticnet) = \left(1 + \frac{\lambda_2}{n}\right)\left\{arg \min_{\beta} \|y - X\beta\|_2^2 + \lambda_2 \|\beta\|_2^2 + \lambda_1 \|\beta\|_1\right\}, \qquad (4)$$

On setting $\alpha = \lambda_2 / (\lambda_1 + \lambda_2)$, the estimator Eqn. (5) is seen to be similar to the minimizer of:
$$\hat{\beta}(elasticnet2) = arg \min_{\beta} \|y - X\beta\|_2^2, subjected\ to\ P_\alpha(\beta) = (1 - \alpha) \|\beta\|_1 + \alpha \|\beta\|_2^2$$
$$\leq s\ for\ some\ s \qquad (5)$$

where $P_\alpha(\beta)$ is the elastic net penalty Eqn. (5). The elastic net clarifies to simple ridge regression when $\alpha = 0$ and to the lasso when $\alpha = 1$. The $\ell_1$ part of the elastic net does self-regulating predictor selection, while the $\ell_2$ part encourages grouped selection and balances the solution paths with respect to random sampling, thereby improving prediction. By activating a categrization effect during predictor selection, like that a bunch of extremely correlated variables likely to have coefficients of uniform magnitude, the elastic net can choose bunches of correlated predictors when the groups are not known in advance. Different from the lasso, when $p \gg n$, the elastic net selects more than $n$ predictors. However, the elastic net needs the oracle property.

## 2.6 Relaxed Lasso

Relaxed lasso (relaxo) is a generalization of the lasso shrinkage technique for linear regression. One predictor selection and other parameter evaluation is attained by regular lasso, still both steps do not certainly employ the identical penalty parameter. The results include all lasso solutions but allow for sparser models while accepting related predictive performance if many predictors are present. A two-phase methd for calculating the relaxed lasso estimator is then projected. The set of predictors selected by the lasso estimator $\hat{\beta}(lasso)\ \lambda$ is represented by $\mathcal{M}_\lambda$,

$$\mathcal{M}_\lambda = \left\{ 1 \le k \le p | \hat{\beta}_k(lasso) \ne 0 \right\}. \tag{6}$$

For sufficiently large penalties $\lambda$ (e.g. for $\lambda > 2max_k n^{-1} \sum_{i=1}^{n} Y_i X_i^k$), the selected model is the null set, $\mathcal{M}_\lambda = \phi$ , as all components of the estimator in Eqn. (2) are identical to zero. In the non arrival of a $\ell_1$- penalty and if the number of predictors $p$ is smaller than the number of observations $n$, all predictors are in general selected, so that $\mathcal{M}_0 = \{1, \ldots \ldots, p\}$ in this case.

The $\ell_1$– penalty for the ordinary lasso-estimator Eqn. (2) has two results , model selection and shrinkage estimation. On the one hand, a valid set of coefficients is set to zero and hence not included from the selected model. On the other hand, for all predictors in the selected model $\mathcal{M}_\lambda$ , coefficients are shrunken towards zero compared to the least-squares solution. These two results are clearly related and can be best understood in the framework of orthogonal design as soft-thresholding of the coefficients. How ever, it is not at once understandable whether it is certainly optimal to control these two results, model selection on the one part and shrinkage evaluation on the other part, by a single parameter only. As an example, it might be adorable in some situations to estimate the coefficients of all selected predictors without shrinkage, according to a hard-thresholding of the coefficients.

As a standard of one soft- and the other hard-thresholding , we control model selection and shrinkage estimation by two discrete parameters $\lambda$ and $\phi$ including the relaxed lasso estimator. The relaxed lasso estimator is expressed for $\lambda \in [0, \infty)$ and $\phi \in (0,1]$ as

$$\hat{\beta}^{\lambda\phi} = arg \min_{\beta} n^{-1} \sum_{i=1}^{n} \left( Y_i - X_i^T \{ \beta \cdot 1_{\mathcal{M}_\lambda} \} \right)^2 + \phi \lambda \|\beta\|_1, \tag{7}$$

where $1_{\mathcal{M}_\lambda}$ is the indicator function on the set of predictors $\mathcal{M}_\lambda \subseteq \{1, \ldots \ldots, p\}$ so that for all $k \in \{1, \ldots \ldots, p\}$,

$$\{ \beta \cdot 1_{\mathcal{M}_\lambda} \}_k = \begin{cases} 0 & k \notin \mathcal{M}_\lambda, \\ \beta_k & k \in \mathcal{M}_\lambda. \end{cases} \tag{8}$$

Note that only predictors in the set $\mathcal{M}_\lambda$ are count for the relaxed lasso estimator. The parameter $\lambda$ controls thus the predictor selection part, as in ordinary lasso estimation. The relaxation parameter $\phi$ controls on the other part the shrinkage of coefficients. If $\phi = 1$, the lasso and relaxed lasso estimators are alike. For $\phi < 1$, the shrinkage of coefficients in the selected model is downsized compared to ordinary lasso estimation. The case of $\phi = 0$ needs special concern, as the definition above would produce a degenerate solution.

A method is developed to compute the exact solutions of the relaxed lasso estimator. The parameters $\lambda$ and $\phi$ can then be selected e.g. by cross-validation. The method  is based on the lars-algorithm [13]. As the relaxed lasso estimator is parameterized by two parameters, a two-dimensional manifold has to be covered to find all solutions.

## 2.7 Least Angle Regression(LARS)

The evolution of least angle regression (lars) [13] , which can quickly be specialized to provide all lasso solutions in a highly efficient usage, denotes a major breakthrough. Lars is a smaller greedy variant of standard forward selection approach such as all subsets, Forward Selection and Backward Elimination. An interesting characteristic of lars is that it implements the lasso by a simple correction . The lars modification computes all possible lasso estimates for a given problem in an order of importance which requires a much smaller amount of computational time then previous methods.

Least angle regression is a formalized version of the Stage wise procedure [13]. Lars  is closely related with lasso, and in fact provides an deeply efficient algorithm for computing the entire lasso path. Lars uses a similar scheme as forward stepwise regression, but only enters "as much" of the predictor as it deserves. At the first step, it identifies the predictor most correlated with the response; fits the predictor completely, lars  moves the coefficient of this predictor continuously toward its least square value, as soon as another predictor "catches up" in terms of correlation with the residual, the process is waited. The second predictor joins the active set, and their coefficients are moved together in a way that keeps their correlations bind and decreasing. This process is continued until all the predictors are in the model and ends at the full least-squares fit. The lars method can be recap as follows:

1. Transformation of the predictors to have mean zero and unit norm. Start with residual
$r = y - \qquad \hat{y}, \beta_1, \beta_2, \ldots \ldots, \beta_p = 0.$

2. Identify the predictor $x_j$ too correlated with $r$.

3. Advance $\beta_j$ from 0 towards its least-squares coefficient $\langle x, r \rangle$ until some other competitor $x_k$ has as  much correlation with the current residual as does $x_j$

4. Advance $\beta_j$ and $\beta_k$ in the direction expressed by their joint least squares coefficient of the current  residual on $(x_j, x_k)$ just before few other competitor $x_l$ has as sufficient correlation with the common  residual.
*lasso correction*. If a non-zero coefficient hits zero, drop its predictor from the active set of predictor and recompute the current joint least squares direction.

5. Carry in this method just before all  $p$ predictors have been filed. After $min(n-1, p)$ steps, we reach  at the full least-squares solution.

If $p > n - 1$ the lars algorithm reaches a zero residual solution after $n - 1$ steps steps (the $-1$ is because we have centered the data).
The *lasso correction*  in the fourth step is an effective method of calculating the result to any lasso problem, particularly when $p \gg n$.

## 2.8 Fitting and analyzing models

The whole path of results (in $\lambda$) for the ridge regression, lasso and elastic net models were calculated using the path wise cyclical coordinate descent algorithms– computationally effective techniques for find out these convex optimization examples– in *glmnet* in R [5]. We used ten-fold cross valdation (CV) within *glmnet* to entirely search for the optimal $\lambda$. The relaxed lasso was fitting using the *relaxo* package in R whereas the least angle regression using an R package *lars*. Similarly lars models selects optimal parameter $\lambda$ using ten-fold Cross Validation. We used ten-fold cross validation within *relaxo.*to search for optimal $\lambda$ and $\phi$. In order to support an overview to the models and the experimental data, visual descriptions were shown for the regularized regression model. A regularized profile plot of the coefficient paths for a five fitted models are shown. Predictive accuracy was also assessed as the root means squared error (RMSE) and coefficient of determination (R square).

## 2.9 Benchmark Experiments

Move in benchmark experiments for comparison of regularization shrunken regression models have been somewhat recent. The experimental performance distributions of a set of regularized linear regression algorithms are estimated, compared, and ordered. The resampling process used in these experiments must be investigate in further detail to determine which method produces the most accurate analysis of model influence. Resampling methods to be compared include cross-validation [21-23]. We can use resampling results to make orderly and inorderly comparisons between models [21-22] Each model performs 25 independent runs on each sub sample and report minimum, median, maximum, mean of each performance measure over the 25 runs.

## 3. RESULTS AND DISCUSSION

Up until this point, we have described several regularized linear regression methods. In this section, we conduct experiments to investigate their individual performances. Regularization techniques can make better the predictive error of the model by lowering the variability in the measures of regression coefficient by shrinking the estimates toward zero. These five methods will shrink some coefficient estimates to exactly 0, thus supplying a scheme of predictor selection. Regularization plots are plots of the regression coefficients vs the regularization penalty $\lambda$. When searching a range of values for the suitable penalty coefficient, it provides a view of how the regression coefficients change over that range. So we can lay the regularization path which demonstrates how the coefficient of each input predictors changes when the $\lambda$ adjusts and choose the suitable $\lambda$ that filter out the number of input predictors for us. Here is the output of these plots.

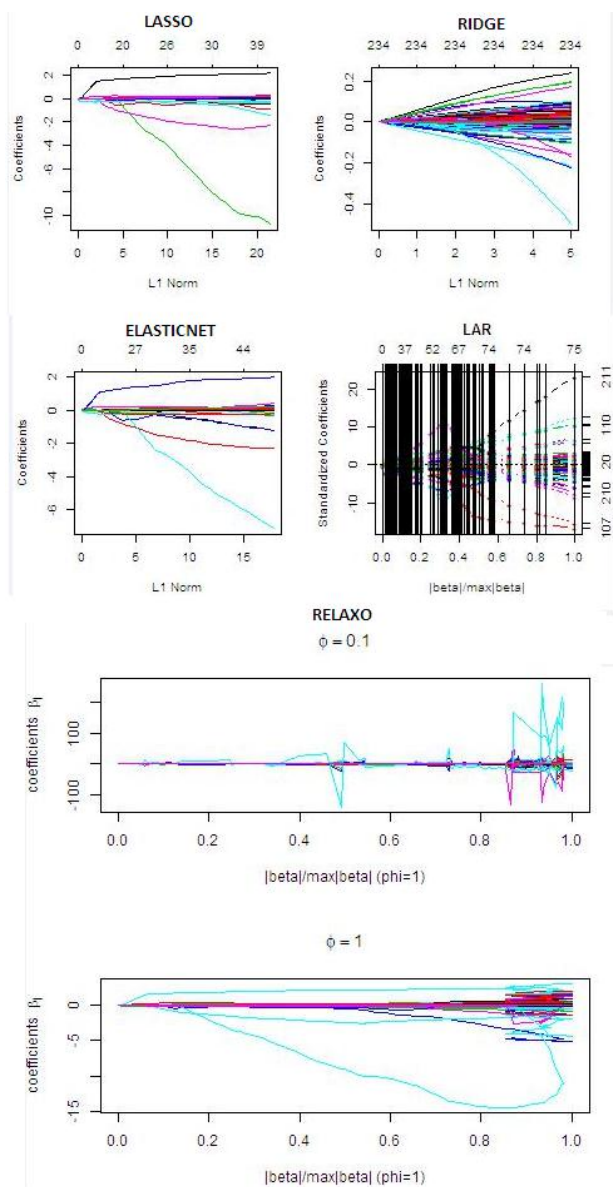

Figure 1. Coefficient paths of Lasso, Ridge, Elasticnet, Lars and Relaxed lasso linear models for molecular descriptors of Oxazolines and Oxazoles derivatives dataset

In Figure 1 these plots shows nonzero model coefficients as a function of the regularization parameter $\lambda$. Because there are 234 predictors and linear models, there are 234 curves. As $\lambda$ increases, A Regularized models showing various coefficients to zero, removing them from the model. The paths for the linear regularized regression models were obtained by finding the transition points. RMSE and R square values were used to compare model prediction accuracies for the lasso, ridge, elasticnet, lars and relaxed lasso regression models. Comparing the resampling performance the effect of prediction of antituberculer activity using Oxazolines and Oxazoles derivatives are demonstrated in Figure 2.

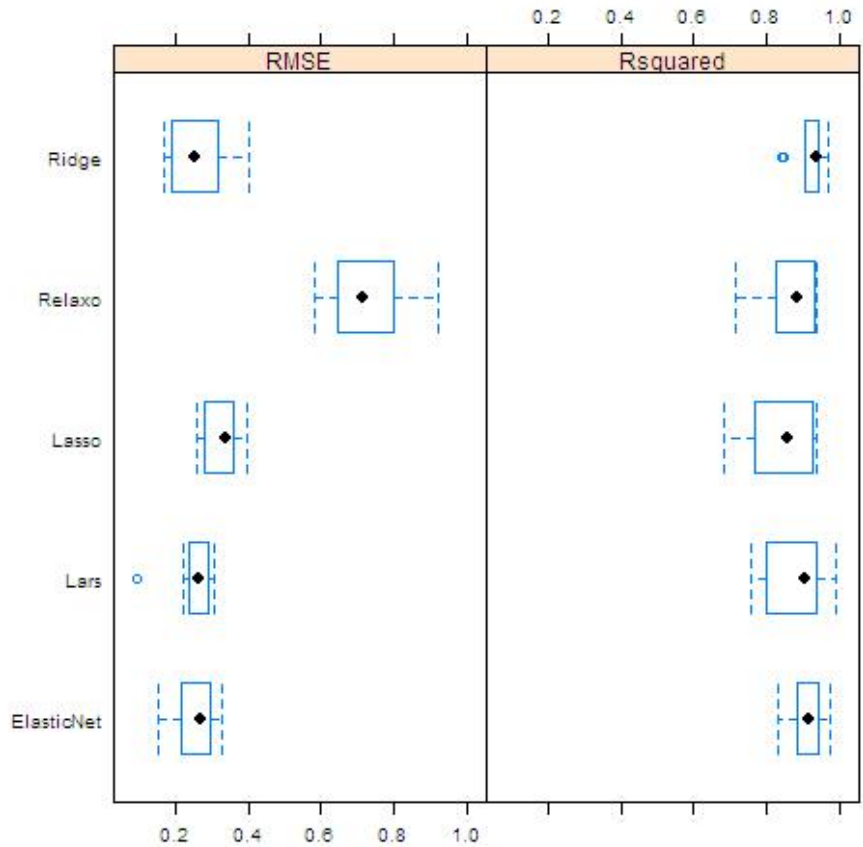

Figure 2. Box-whisker plot of the RMSE and R squared values for Regularized linear models. The elasticnet model, lars model and the ridge model give the smallest prediction errors; the lars yields the smallest RMSEP spread and ridge yields the smallest R Square spread.

The RMSE and R Square values for the five models for prediction of antitubercular activity are comparable. as shown in Figure 2. Lars , lasso and elasticnet appear to have slightly smaller RMSE error spreads  than relaxed lasso and ridge. Ridge, elasticnet and relaxed lasso appear to have slightly smaller R square error spreads than lasso and lars. Pair-wise comparisons of model RMSE and R square values using Student's $t$-test reveal that there is statistical difference in the prediction accuracies of the five regularized models. These results are shown in Table 1, which gives both the $p$-values and the absolute differences in RMSE and R square for the model comparisons. None of the $p$-values are smaller than the specified significance level α = 0.05. The null hypothesis is not dropped; in the case of this data set, there is no statistically meaningful difference in performance among these five regression methods.

Table 1. Pair-wise comparisons of RMSE and R Square differences and $p$-values

| RMSE differences (upper diagonal) and $p$-values (lower diagonal) | | | | | |
|---|---|---|---|---|---|
| | Lasso | Elasticnet | Lars | Relaxo | Ridge |
| Lasso | | 0.067958 | 0.074415 | -0.395483 | 0.069507 |
| Elasticnet | 0.3885 | | 0.006456 | -0.463441 | 0.001549 |
| Lars | 0.3208 | 1.0000 | | -0.469898 | -0.004908 |
| Relaxo | 6.873e-06 | 3.593e-05 | 1.829e-05 | | 0.464990 |
| Ridge | 0.6333 | 1.0000 | 1.0000 | 6.733e-05 | |
| R Square differences (upper diagonal) and $p$-values (lower diagonal) | | | | | |
| | Lasso | Elasticnet | Lars | Relaxo | Ridge |
| Lasso | | -0.06290 | -0.03779 | -0.02468 | -0.07643 |
| Elasticnet | 1.0000 | | 0.02511 | 0.03822 | -0.01352 |
| Lars | 1.0000 | 1.0000 | | 0.01311 | -0.03863 |
| Relaxo | 1.0000 | 1.0000 | 1.0000 | | -0.05175 |
| Ridge | 0.4211 | 1.0000 | 1.0000 | 1.0000 | |

It should be observed that the $p$-value for this pair-wise comparison is 6.873e-06 (Table 1) for RMSE, which is not valid at α = 0.05, but it is still a much smaller $p$-value than those obtained for the other four pair-wise comparisons. To test for pair-wise differences, we use Tukey differences.

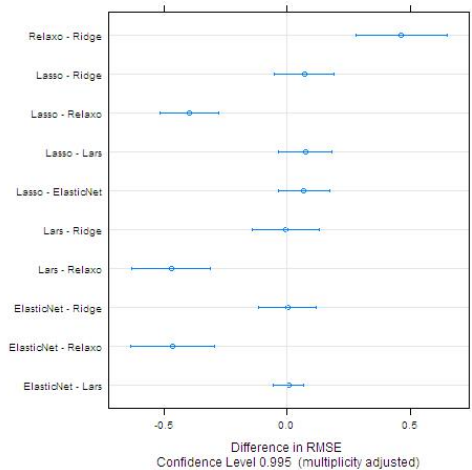

Figure 3. Asymptotic simultaneous confidence sets for Tukey all-pair regularized linear regression models comparisons of the RMSE errors after alignment.

As a major advantage compared to the non-parametric methods, can calculate simultaneous confidence intervals. Figure 3 shows the corresponding 99% model wise confidence intervals where the bars of a given comparison are outside the 0 difference in RMSE line there is a statistically significant difference at the 99% level present. The blue dot indicates the estimated magnitude of this difference. The differences between (lars, ridge), (elasticnet, ridge) and (elasticnet, lars) are not significant, the corresponding confidence intervals intersect zero and overlap each other.

Table 2. Accuracy of predictions of the five regularized linear models

|  | RMSE | | R square | |
|---|---|---|---|---|
|  | **Train** | **Test** | **Train** | **Test** |
| Lasso | 0.4627773 | 0.391653 | 0.8114705 | 0.9050635 |
| Ridge | 0.1164121 | 0.2773841 | 0.9835022 | 0.8689849 |
| ElasticNet | 0.1656584 | 0.306454 | 0.9577759 | 0.8477245 |
| Lars | 0.1752777 | 0.2878424 | 0.952097 | 0.8703903 |
| Relaxo | 0.8552319 | 0.882294 | 0.7891734 | 0.8987195 |

In our study, regularized regression methods to predict antitubercular activity of Oxazolines and Oxazoles derivatives. Predictive accuracy of all five regularized model evaluated as the Square of usual correlation between predictor and response (mean inhibitory concentration) i.e R square and the root mean squared error (RMSE). RMSE provides baseline measure of predictive accuracy. In this case descriptor dataset is splits into training set and test set. Training set comprises seventy six observations and test set comprises twenty four observations.

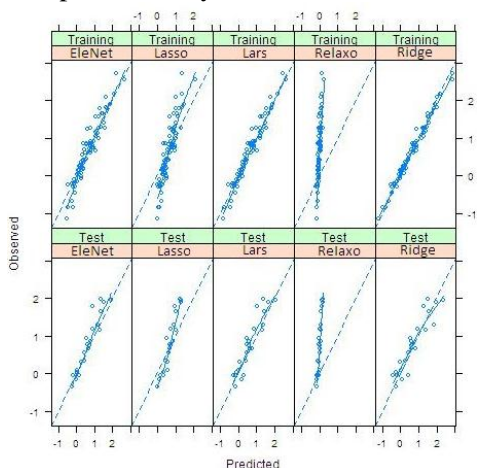

Figure 4. Comparison of prediction performance of trained and tested models obtained by five regularized linear methods for Oxazolines and Oxazoles derivatives descriptor dataset.

All results reported are for the train set and test set. The predictive estimation results are summarized in Table 2. The obtained both RMSE and R square values of trained ridge model are 0.1164121 and 0.9835022 comparatively better than elasticnet and lars models. Rest of the trained lasso and relaxed lasso models shows low performance than elasticnet and lars models. The obtained RMSE value for tested ridge model is 0.2773841 comparatively better than lars and elasticnet models as well as obtained R square value for tested lasso model is 0.9050635 comparatively better than relaxed lasso and lars. Overall ridge model exhibits good predictive accuracy. Figure 4 shows the performance comparison of the five methods for antitubercular activity prediction.

## 4. CONCLUSIONS

In this paper, we studied and examined the performance of the five regularized linear regression models. We presented characteristics of regularized methods through regularized profile plots as well as we presented exploratory and inferential analyses of benchmark experiments. Benchmark experiments show that this method is the primary choice to evaluate learning algorithms. It should be observed that the scheme can be used to compare a set of algorithms but does not propose a model selection. The results for the regularized regression suggest that we may observe performance differences with barely high power. We have compared the predictive accuracies with all five models among ridge model exhibits better overall predictive performance.

## ACKNOLDGEMENTS

We thankfully to the Department of Computer Science Mangalore University, Mangalore India for technical support of this research.

## Authors

**Doreswamy** received B.Sc degree in Computer Science and M.Sc Degree in Computer Science from University of Mysore in 1993 and 1995 respectively. Ph.D degree in Computer Science from Mangalore University in the year 2007. After completion of his Post-Graduation Degree, he subsequently joined and served as Lecturer in Computer Science at St. Joseph's College, Bangalore from 1996-1999.Then he has elevated to the position Reader in Computer Science at Mangalore University in year 2003. He was the Chairman of the Department of Post-Graduate Studies and research in computer science from 2003-2005 and from 2009-2008 and served at varies capacities in Mangalore University at present he is the Chairman of Board of Studies and Professor in Computer Science of Mangalore University. His areas of Research interests include Data Mining and Knowledge Discovery, Artificial Intelligence and Expert Systems, Bioinformatics ,Molecular modelling and simulation ,Computational Intelligence ,Nanotechnology, Image Processing and Pattern recognition. He has been granted a Major Research project entitled "Scientific Knowledge Discovery Systems (SKDS) for Advanced Engineering Materials Design Applications"from the funding agency University Grant Commission, New Delhi,India.He has been published about 30 contributed peer reviewed Papers at national/International Journal and Conferences.He received SHIKSHA RATTAN PURASKAR for his outstanding achievements in the year 2009 and RASTRIYA VIDYA SARASWATHI AWARD for outstanding achievement in chosen field of activity in the year 2010.

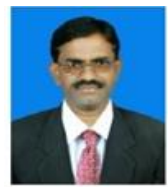

**Chanabasayya .M. Vastrad** received B.E. degree and M.Tech. degree in the year 2001 and 2006 respectively. Currently working towards his Ph.D Degree in Computer Science and Technology under the guidance of Dr. Doreswamy in the Department of Post-Graduate Studies and Research in Computer Science , Mangalore University

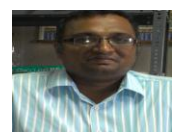